\documentclass[10pt,twocolumn,letterpaper]{article}

\usepackage{wacv}
\usepackage{times}
\usepackage{epsfig}
\usepackage{graphicx}
\usepackage{amsmath}
\usepackage{amssymb}
\usepackage{siunitx}
\usepackage{array}
\usepackage{booktabs}

\def\wacvPaperID{1} 

\wacvfinalcopy 

\ifwacvfinal
\def\assignedStartPage{9876} 
\fi

\ifwacvfinal
\usepackage[breaklinks=true,bookmarks=false]{hyperref}
\else
\usepackage[pagebackref=true,breaklinks=true,colorlinks,bookmarks=false]{hyperref}
\fi
\ifwacvfinal
\else
\fi

\graphicspath{{images/}} 
\usepackage{subcaption}
\newcommand{\figref}[1]{\figurename~\ref{#1}}

\begin{document}
\raggedbottom

\title{
    ShineOn: Illuminating Design Choices for \\ Practical Video-based Virtual Clothing Try-on
}

\author{
    Gaurav Kuppa\ensuremath{^{1}}\\
    San Jose State University\\
    {\tt\small gaurav.kuppa@sjsu.edu}
    \and
    Andrew Jong\ensuremath{^{1}}\\
    San Jose State University\\
    Carnegie Mellon University\\
    {\tt\small ajong@andrew.cmu.edu}
    \and
    Xin Liu\\
    Nanyang Technological University\\
    {\tt\small veralau9425@gmail.com}
    \and
    Ziwei Liu\\
    Nanyang Technological University\\
    {\tt\small ziwei.liu@ntu.edu.sg}
    \and
    Teng-Sheng Moh\\
    San Jose State University\\
    {\tt\small teng.moh@sjsu.edu}
}

\maketitle
\begin{abstract}
\footnotetext[1]{These authors equally contributed to this paper.}
Virtual try-on has garnered interest as a neural rendering benchmark task to evaluate complex object transfer and scene composition.
Recent works in virtual clothing try-on feature a plethora of possible architectural and data representation choices.
However, they present little clarity on quantifying the isolated visual effect of each choice, nor do they specify the hyperparameter details that are key
to experimental reproduction.
Our work, \emph{ShineOn}, approaches the try-on task from a bottom-up approach and aims to shine light on the visual and quantitative effects of each experiment. 
We build a series of scientific experiments to isolate effective design choices in video synthesis for virtual clothing try-on.
Specifically, we investigate the effect of different pose annotations, self-attention layer placement, and activation functions on the quantitative and qualitative performance of video virtual try-on. 
We find that DensePose annotations not only enhance face details but also decrease memory usage and training time.
Next, we find that attention layers improve face and neck quality.
Finally, we show that GELU and ReLU activation functions are the most effective in our experiments despite the appeal of newer activations such as Swish and Sine.
We will release a well-organized code base, hyperparameters, and model checkpoints to support the reproducibility of our results.
We expect our extensive experiments and code to greatly inform future design choices in video virtual try-on. 
Our code may be accessed at \url{https://github.com/andrewjong/ShineOn-Virtual-Tryon}.
\end{abstract}


\section{Introduction}\label{sec:introduction}

The age of the internet has led to an unprecedented consumer shift to online marketplaces.
With the convenience and wide selection provided by online stores, shopping is easier than ever.
However, purchasing clothes online remains a difficult proposition because end-consumers cannot accurately judge clothing fit and appearance on themselves.
Constant returns of ill-fitted, unflattering clothing negatively impact the environment through wasted shipping and packaging.
Fortunately, this problem has the potential to be alleviated with emerging virtual try-on technology. Virtual try-on aims to let users quickly visualize themselves in different outfits through a camera and digital display.
Being able to achieve a high-quality, real-time fashion virtual try-on system may boost online retail sales, as well as cut down on the carbon footprint produced by packaging and returns.

Beyond the product application, the academic neural rendering community also 
has a vested interest in virtual try-on, as it serves as a benchmark task to evaluate complex object transfer and scene composition.
This complexity is embodied in the criteria to accurately maintain the user's identity, render the cloth product at the appropriate location, follow the user's body proportions, preserve cloth texture detail, exhibit smooth temporal dynamics, establish temporal consistency, and blend well with the scene's lighting.

Many virtual try-on works that emerged in the past three years explore deep learning methods. 
These mostly focus on image try-on \cite{wang2018characteristicpreserving, wu2018m2etry, dong2019towards, Yang_2020_CVPR, Neuberger_2020_CVPR}, and only recently has there been an exploration of video-based try-on \cite{Jong2019ShortVideos, Dong2019FWGAN}.

There has been substantial work towards virtual try-on, including image-image translation, perceptual loss, and the improvement of human parsing techniques.
Image-image translation \cite{isola2018imagetoimage} serves to pivotal in transferring a cloth to a target image.
Common losses like perceptual loss \cite{7780634, johnson2016perceptual} are highly effective in retaining quality transfer.
The proliferation of fashion datasets \cite{liu2016mvc, liu2016deepfashion, ge2019deepfashion2, gong2017look} and human parsing techniques such as cloth segmentation, body segmentation \cite{liu2016deepfashion, ge2019deepfashion2}, pose annotation \cite{guler2018densepose}, and more have increasingly improved in quality. These techniques are critical for robust virtual try-on.

We want to develop a deep, scientific understanding of video virtual try-on and display transparency in our methods to enable more great work in the field of neural rendering, and specifically virtual try-on. We aim to shed light on the workings of virtual try-on and lead to meaningful insights on how to improve it. In particular, we test the effectiveness of DensePose pose annotations, self-attention, activation functions, and optical flow. We accumulate the results of our studies and compare our resulting ShineOn approach with existing try-on methods.

The specific contributions of this paper are (1) we present transparent, comprehensive bottom-up experiments testing pose annotations, self-attention, activation functions, and optical flow, (2) demonstrate a decrease in memory usage, training time, and improved face detail transfer by using DensePose pose annotations rather than CocoPose pose annotations, and (3) propose \emph{ShineOn}, the accumulation of our most effective methods for practical video virtual try-on.
At the time of writing, ShineOn is the only public video virtual try-on codebase readily available for scientific verification and reproducibility, which is critical for the future of try-on works.


\section{Related Works}

\subsection{Image Virtual Try-on}

There has been a large proliferation of publicly available datasets \cite{liu2016mvc, liu2016deepfashion, ge2019deepfashion2, gong2017look} that have pushed forward the multiple-component tasks of try-on mechanics. These datasets are of great importance to advanced tasks, such as human parsing, cloth segmentation, body segmentation, pose estimation. These annotations led to more robust human parsing models, such as SSL and JPPNet \cite{gong2017look, liang2018look}.

Quite a few proposed network architectures have been developed to accomplish body and cloth segmentation \cite{liu2016deepfashion, ge2019deepfashion2}. Pose information has been embedded through CocoPose, and more recently through DensePose \cite{guler2018densepose}.

Realistic virtual try-on requires transferring of high-level structures such as cloth pattern, design, text, and texture. In addition to L1 loss, perceptual and feature losses \cite{7780634, johnson2016perceptual, zhang2018perceptual, ding2020image} consist of important components to address this issue.

The history of virtual try-on methods and deep learning dates back to 2017.
These methods are dependent on effective and generalizable human parsing methods. From Jetchev and Bergmann's Conditional Analogy GAN \cite{jetchev2017conditional}, virtual try-on has seen significant growth in the years \cite{Han_2018_CVPR, wang2018characteristicpreserving, raj2018swapnet, 8636265, wu2018m2etry, dong2019towards}. These methods were iteratively improved by introducing perceptual loss, removing adversarial loss, and experiment with different network architectures to synthesize more detailed virtual try-on outputs. The general architecture for these methods is a two-stage approach that involves cloth warping and person rendering. Some virtual try-on networks employ more than two stages, which we found unnecessary. A two-stage approach has enough expressiveness for the model to learn how to do virtual try-on.


\subsection{Video Virtual Try-on}

Given the significant development for virtual try-on with images, the next natural step is to investigate virtual try-on for video. Video try-on would allow a user to easily examine the clothing’s appearance on their body at multiple angles, instead of processing individual images. However, video try-on raises new challenges, such as how to handle temporal consistency \cite{lai2018learning, zhang2019exploiting} between video frames.

FW-GAN \cite{Dong2019FWGAN} requires a video of a reference person wearing the desired clothes. This was collected in a new dataset, VVT \cite{Dong2019FWGAN}, which was obtained from scraping fashion walk videos on fashion websites. The image of the user is then warped to match the pose of the reference, and the desired clothes from the reference are composited with the warped user.


\subsection{Attention and Activations Functions}
Attention and transformers \cite{xu2015show, yang2016hierarchical, vaswani2017attention} have shown to work plenty in achieving world-class performance in the machine translation task. The notion of self-attention demonstrated the ability to model long-range dependencies in an interpretable way. Image transformers \cite{parmar2018image} and self-attention for convolution neural networks and generative adversarial networks \cite{cordonnier2019relationship, carion2020end} began to popularize. It gave way to self-attention in image-image translation. However, self-attention has not been used for the virtual try-on task yet, our work will verify its power to model the longer-range dependencies when transferring the garment to the model person.

ReLU networks were suggested to have a bias towards learning low-frequency information \cite{rahaman2019spectral}. Similarly, literature \cite{sitzmann2020implicit, xie2020smooth, sitzmann2020implicit, tancik2020fourier} show that smoother activation functions are more effective at achieving robust results for representing and reconstructing media.

\section{Problem Formulation and Challenges}
\begin{figure*}[ht!]
    \centering
    \includegraphics[width=14cm]{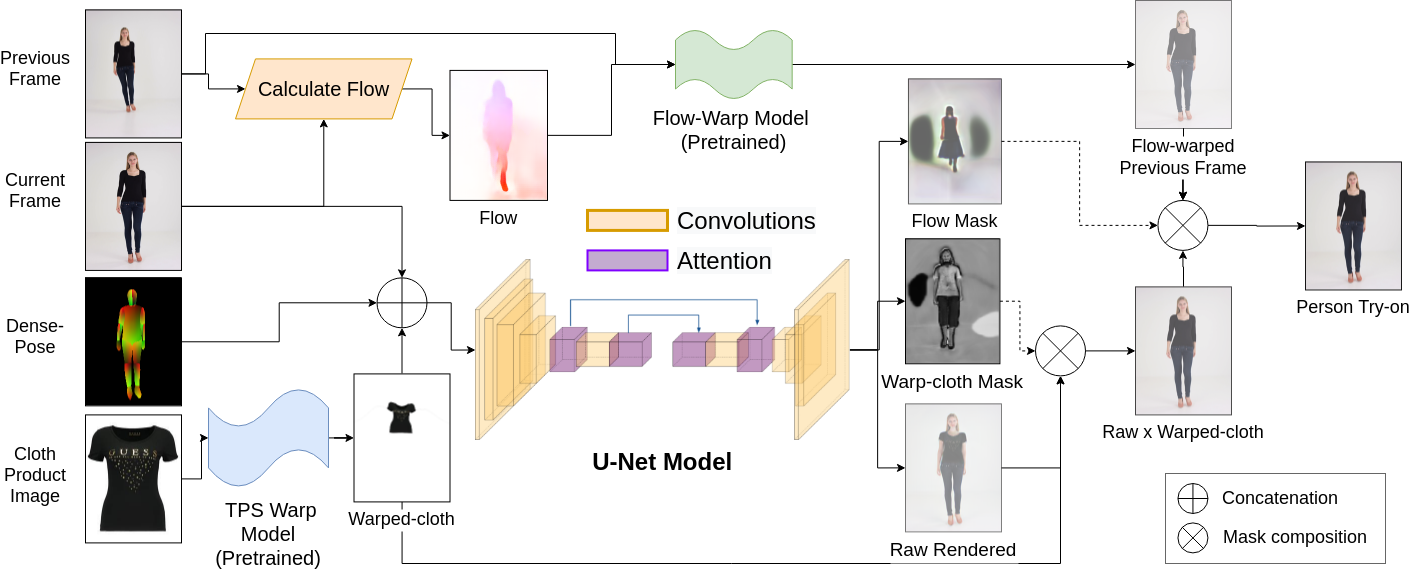}
    \caption{\textbf{Visualization of ShineOn Architecture.} 
    The input person representation $p$ and warped cloth $w$ are fed into the U-Net model with self-attention. The output of the U-Net, after masking, is the person and warped cloth composed together. For experiments with Flow, we add the top branch to finally compose with a flow-warped previous frame.}
    \label{fig:architecture}
\end{figure*}
\subsection{Problem Formulation}\label{subsec:problem_formulation}

\begin{figure}[h]
    \centering
    \includegraphics[width=8cm]{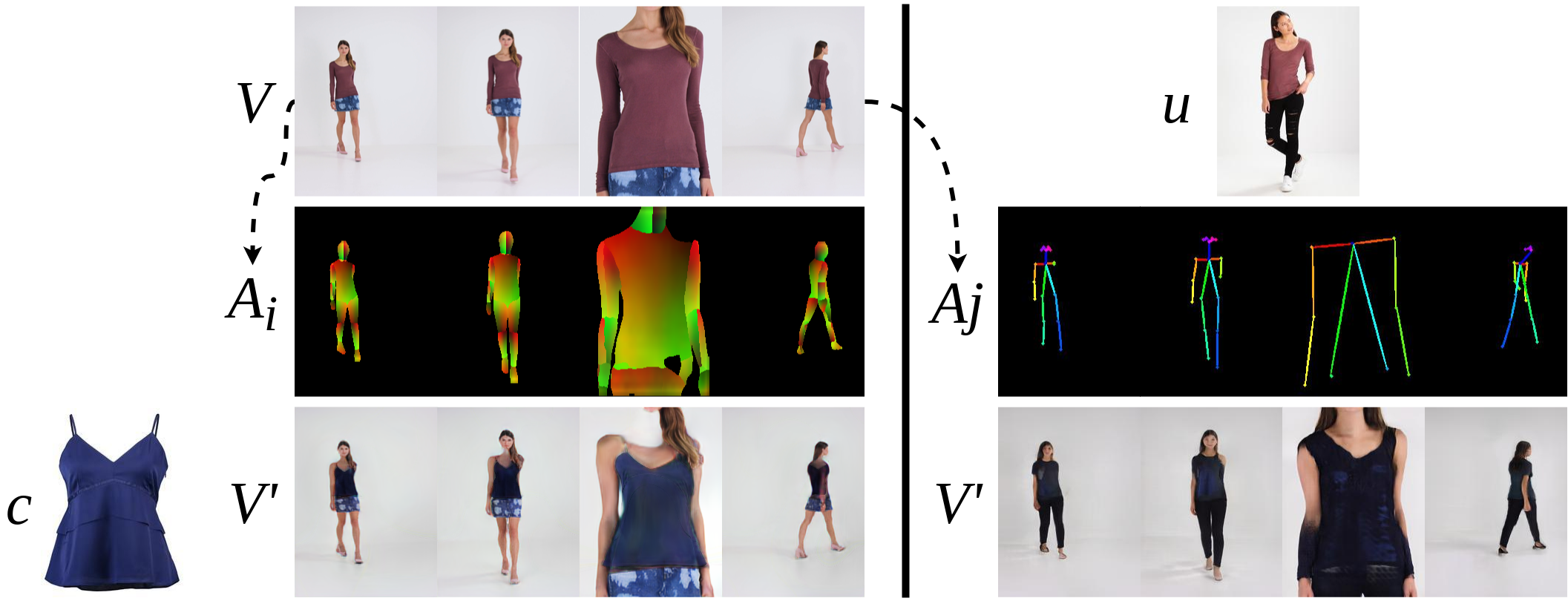}
    \caption{
        Problem formulation of \emph{ShineOn} (ours, left) compared to \emph{FW-GAN}'s \cite{Dong2019FWGAN} (right). 
        Dashed arrows show that the annotations $A_t$ are parsed from the source frames in $V$.
        The key difference is that our formulation directly treats the video as the user input, 
        whereas \cite{Dong2019FWGAN} uses a separate still user image and reposes the user to the video's key points. 
        This means the compared results feature different users (Left: blonde hair, Caucasian, and blue shorts. Right: black hair, Asian, and black jeans).
    }
    \label{fig:problem_formulation}
\end{figure}

\noindent\textbf{Try-on Task for Inference.}
Given 
a desired cloth product image $c$, 
a video of a user with $n$ frames $V (v_1, …, v_n)$, 
video annotation type $a_{t}\in{A}$, and 
each annotation type generated for every frame $A(a_{t_1}, …, a_{t_n})$, 
we develop a model to synthesize a new video $V’$, in which the user from video $V$ is realistically wearing cloth $c$. See \figref{fig:problem_formulation} for details.

It is important to note that our formulation is fundamentally different that that posed in our closest related work, \emph{FW-GAN} \cite{Dong2019FWGAN}.
FW-GAN proposes to input only a single user image $u$ instead of inputting a video of the user. 
The output $V'$ then synthesizes the user to follow an arbitrary keypoint pose sequence that is parsed from a separate video. See \figref{fig:problem_formulation} for this comparison.

We argue that our formulation is more practical from an applications standpoint.
In application, a user may want to see the clothes at different angles on their body and move their bodies as they wish to instantaneously change the view.
When applied in real-time, this level of control corresponds to looking in a real-life mirror.
We adopt this in our approach because it permits direct user control, rather than reposing a still user image to arbitrary keypoints.
This formulation involving direct control lets us assume that video frames of the user are available. 

In either formulation, the synthesized video $V’$ aims to meet the criteria described in Section \ref{sec:introduction}.

\noindent\textbf{Reconstruction Task for Training.}
It is challenging to collect large amounts of data with ground truth try-on pairs.
Such ground truth data would require the same person to hold the exact same pose across two or more different outfits. 
This is already a challenge for a single person, much less a whole dataset, as people naturally to try on different clothes.
Therefore, training for virtual try-on is often framed as a self-supervised task.

We follow the same self-supervised reconstruction task as our base experiment \emph{CP-VTON} \cite{wang2018characteristicpreserving}. 
In the reconstruction task, the target garment that the user is wearing (e.g. shirt) in the training video is masked out. 
We then train the model to synthesize the missing garment in each frame given its corresponding isolated cloth product image $c$. This way we may calculate a loss between the synthesized and ground truth train sample. We use several common losses, $\mathcal{L}_{L1}$, $\mathcal{L}_{mask}$, and $\mathcal{L}_{VGG}$, for the reconstruction task. As stated by Pix2Pix \cite{isola2018imagetoimage}, $\mathcal{L}_{L1}$ is essential to achieve image translation, and is the reason that many modern virtual try-on methods use these core components. Additionally, we utilize $\mathcal{L}_{mask}$ to retain characteristic details of cloth and alignment with the target person's body shape. Lastly, realistic virtual try-on requires transferring of high-level structures such as cloth pattern, design, text, and texture; $\mathcal{L}_{VGG}$ is a vital component to address this issue.

\subsection{Dataset}
We use the \emph{Video Virtual Try-on} (VVT) dataset supplied by \cite{Dong2019FWGAN} for our video virtual try on task.
The VVT dataset contains 791 videos recorded at 30 frames-per-second at 192$\times$256 resolution, and each video has a duration between eight and ten seconds.
This results in roughly 250-300 frames per video.
The train and test set contain 159,170 and 30,931 frames respectively (equivalent to 84\% and 16\% of the total). 
Each video has a corresponding isolated cloth product image. The VVT dataset only contains product images of upper clothes (shirts, blouses, sweaters, etc).

Because the final evaluation task of try-on is distinct from the reconstruction training task, we choose to treat the given test set as validation for each of our successive experiments. We then examine try-on as the true test in our final evaluation in Section \ref{sec:results} after all design choices are finalized.

\subsection{Architecture}
Our experiments start from the sequential two-module architecture proposed by our baseline \cite{wang2018characteristicpreserving}. 
The first module, \emph{Warp}, warps the cloth product image $c$ to the body shape and pose of the user at each video frame in $V$.
The second module, \emph{Try-on Composition}, produces both a raw synthesized try-on image and a mask. It uses the mask to compose the raw try-on with the previously warped cloth to produce the final try-on result. We refer the reader to \cite{wang2018characteristicpreserving} for details of the base architecture, and to \figref{fig:architecture} for details of how we modify the architecture in our experiments.

Following \cite{Dong2019FWGAN}, we use a pre-trained Warp module. 
We focus on improving the Try-on module as we observed it was responsible for textural artifacts in preliminary experiments.

U-Net \cite{ronneberger2015unet} adds skip connections between corresponding encoder and decoder layers by concatenating their respective outputs. More generally, U-Net has shown to render smoothly synthesized images. This makes the U-Net architecture a good fit for image and video virtual try-on.

The U-Net model outputs a rendered person $p$, and composition mask $m$. The person try-on, $\hat{p}$, is obtained as shown below:

\begin{equation}
	\hat{p} = w * m + p * (1 - m)
\end{equation}

where $w$ is the warped cloth.

\subsection{Challenges}
The U-Net architecture is effective in doing image virtual try-on. It is able to synthesize a cloth onto a single person. There are significant issues with generalizing image virtual try-on to video virtual try-on. We detail several challenges with these methods, that we address with our experiments.

The purpose of the person-representation is to feed in information to the neural network to learn the most effective way to render a person’s image with the warped cloth. Current methods of person representation involve using a large tensor of images including 18-keypoint pose-annotation, body shape image, and reserved regions annotations. These methods depend on this large person representation to aid the network's learning.

The most significant issue of video virtual try-on is the lack of high-quality cloth transfer while retaining strong texture features of the cloth. Methods can work on trained datasets, but fail to generalize at test time. At test time, there is a significant lack of detail. Virtual try-on suffers from flickering details and cloth texture distortion.

\section{Experiments}

We detail several experiments that we created, to judge the efficacy and impact on video virtual try-on. The nature of our experiments is that we test one experimental variable, while not changing other variables. At each experiment, we determine the result of each experiment and recommend the most effective variable as a part of our design choices in ShineOn.

\subsection{Training Setup}
In all experiments, we train the U-Net model to 10 epochs with an accumulated batch size of 64. We use the Adam optimizer \cite{kingma2017adam} with an initial learning rate of 0.0001 that decays linearly after 5 epochs. Additionally, we utilize 16-bit precision training to increase training speed and reduce GPU memory consumption.

\subsection{DensePose}\label{exp:densepose}

\begin{figure}[h]
    \centering
    \includegraphics[width=6cm]{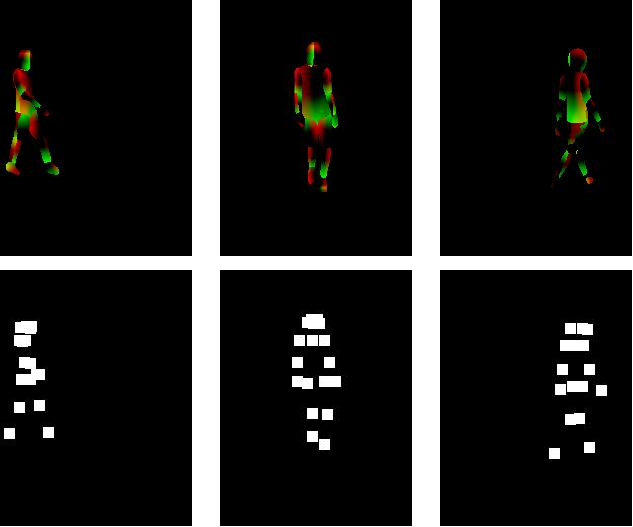}
    
    \caption{\textbf{Visual Comparison between DensePose and CocoPose} DensePose annotations (top row) contain dense 3D body information in the form of UV coordinates, whereas CocoPose annotations (bottom row) are sparse and limited to 2D keypoints.}
    \label{fig:cocopose_densepose}
\end{figure}

DensePose has been shown to model 3D body information in the UV field, and it is an effective representation of body shape, body parts, and includes additional information. We found it beneficial to replace the 18-keypoint CocoPose pose coordinates with the DensePose pose annotation, which show to be just as accurate for pose information. The visual comparison between DensePose and CocoPose, as shown in \figref{fig:cocopose_densepose}, illustrates the increased detail of information compared to the larger CocoPose pose annotation. The benefit of DensePose is that it decreases the size of pose representation by 6 times. Since data loading is the bottleneck of the training pipeline, this decrease significantly decreases the training time of virtual try-on networks.

The 18-keypoint pose representation is a sparse representation of pose information and severely slows down training. We show a 6 times decrease in size of pose-annotation by using DensePose \cite{guler2018densepose} to represent pose-information. Additionally, we also show that DensePose retains far more face details than using CocoPose as a pose annotation. \figref{fig:cocopose_densepose_plot} provides clear quantitative analysis of the effects of using DensePose annotations.

\subsection{Self-Attention}\label{exp:attention}

\begin{figure}[h]
    \centering
    \includegraphics[width=8cm]{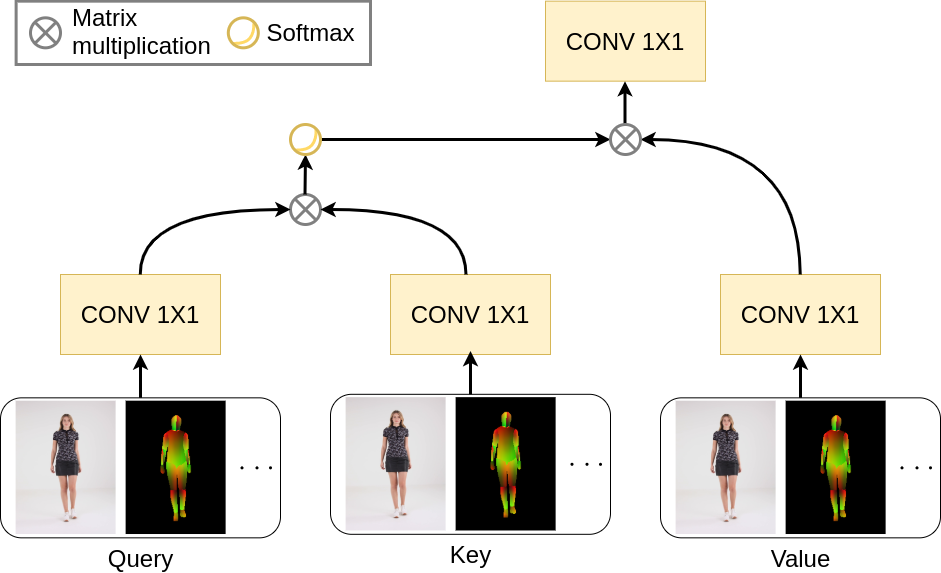}
    \caption{\textbf{Self-Attention Layer for Video Virtual Try-on} Self-attention layers is implemented by feeding in the input representation $i$ or a convolved feature map derived from $i$. Self-attention layers have been shown to model long-range dependencies, and attend to important spatial regions \cite{zhang2019self}.}
    \label{fig:self_attn}
\end{figure}

Self-attention has been repeatedly used to model long-range dependencies, express interpretable results, and more recently, attend to visual and spatial regions of importance. There have been object detection, image generation, and many more computer vision tasks that use self-attention to achieve state-of-the-art performance, respectively.

Within our UNet model, we utilize self-attention layers within our encoder-decoder architecture. The self-attention layers are strategically placed where the input feature map depths are greatest so that the attention module has the capacity to learn the most important features to attend to. Self-attention is computed as:

\begin{equation}
    f(q, k, v) = softmax(qk)’*v
\end{equation}
where q, k and v are query, key and value, respectively.

\figref{fig:architecture} shows our application of self-attention to virtual try-on. We feed in a person representation $p$, or a downsampled representation of $p$.

As shown in \figref{fig:attn_plot}, self-attention does not enable the network to retain more detail of the transferred cloth. There is an improvement of stronger features of and neck details. As opposed to some recommendations \cite{zhang2019self}, we use self-attention where the feature map depth is greatest. This design choice retains similar quality for virtual try-on and reduces the size of the model.
\begin{figure}[bp!]
    \centering
    \includegraphics[width=7.5cm]{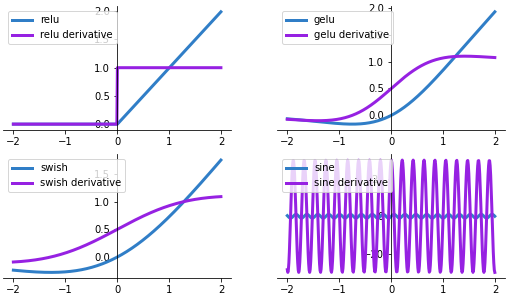}
    \caption{\textbf{Activation Functions and Their Derivatives} From left to right and top to bottom, the activation functions are ReLU, GELU, Swish, and Sine. These activation functions and their derivatives visualize the discontinuity of the ReLU function's gradient, smooth nature of GELU and Swish activation function and their gradients, and erratic nature of Sine function's gradient.}
    \label{fig:activation_functions}
\end{figure}

\subsection{Frequency Robustness}\label{exp:frequency}

\begin{figure*}[h]
    \begin{subfigure}{.25\textwidth}
      \centering
      \includegraphics[width=4cm]{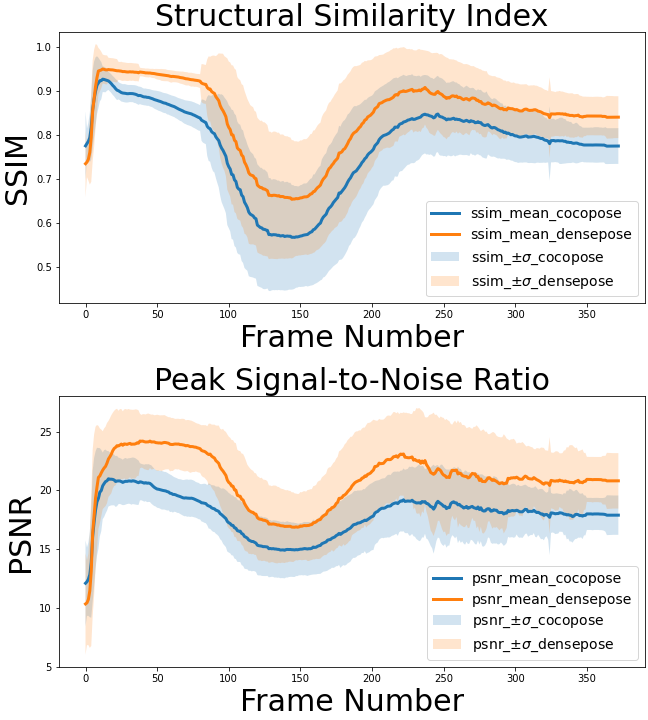}
      \caption{Effect of DensePose}
      \label{fig:cocopose_densepose_plot}
    \end{subfigure}%
    \begin{subfigure}{.25\textwidth}
      \centering
      \includegraphics[width=4cm]{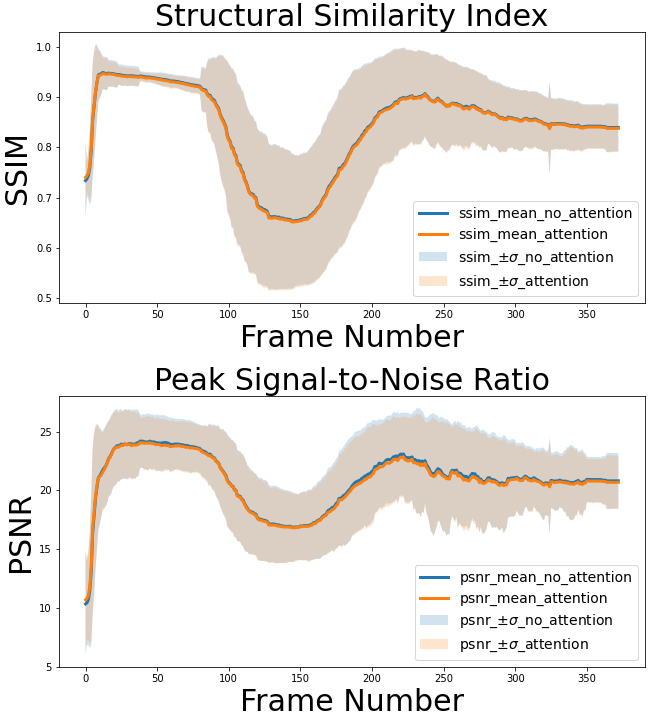}
      \caption{Effect of Attention}
      \label{fig:attn_plot}
    \end{subfigure}%
    \begin{subfigure}{.25\textwidth}
      \centering
      \includegraphics[width=4cm]{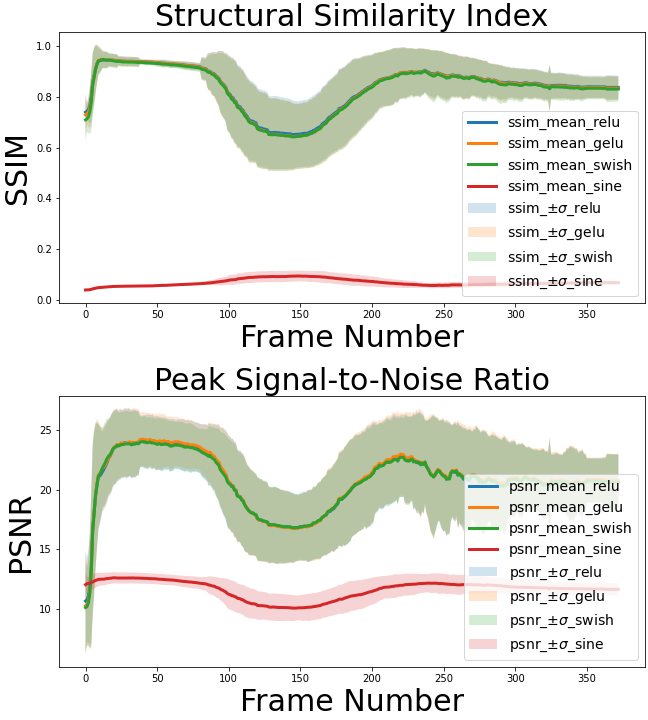}
      \caption{Effect of Activation Functions}
      \label{fig:activation}
    \end{subfigure}%
    \begin{subfigure}{.25\textwidth}
      \centering
      \includegraphics[width=4cm]{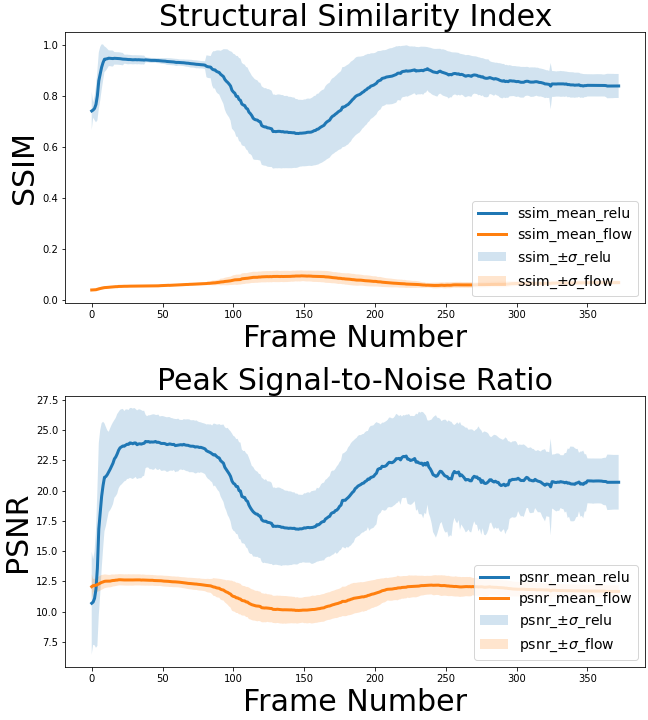}
      \caption{Effect of Optical Flow}
      \label{fig:flow}
    \end{subfigure}%
    \caption{\textbf{Quantitative Comparisons of DensePose, Self-Attention, Activation Functions, and Optical Flow.} We show the mean and standard deviation of metrics across all videos in the VVT dataset for the reconstruction task. We observe that DensePose significantly improves performance over CocoPose. Self-Attention and GELU activation function  improve performance over their alternatives. Lastly, Optical Flow causes video virtual try-on to be worse.}
    \label{fig:activation_flow}
\end{figure*}

Recent literature suggests that ReLU networks are ineffective at rendering and recovering images with high-frequency information. In the garment transfer and virtual try-on domain, preserving high-frequency information corresponds to preserving clothing details and texture information. SIREN and Fourier Features \cite{tancik2020fourier, sitzmann2020implicit} use sine annotations and activation functions, respectively, to retain high-frequency information. Smooth Adversarial Learning \cite{xie2020smooth} shows that smooth gradients for activation functions will lead to more robust models. We hypothesize that smoother activation functions, as shown in \figref{fig:activation_functions}, will lead to more high-frequency information being transferred to the target person.

In this paper, we test several activation functions, ReLU, GELU, Swish, and Sine, \cite{ramachandran2017searching, NIPS2012_4824, hendrycks2020gaussian, tancik2020fourier}, to see the effect on virtual try-on video synthesis. Through our experiments, we determine that ReLU, GELU, and Swish perform similarly on quantitative metrics as per \figref{fig:activation}. However, qualitative inspection reveals that GELU synthesizes more accurate face and neck features. Contrary to recent literature \cite{rahaman2019spectral}, ReLU also synthesizes similar quality results. On the other hand, the Sine activation function performs poorly in the virtual try-on task.

\subsection{Optical Flow}\label{exp:optical-flow}
To improve temporal consistency, we take inspiration from \cite{wang2018vid2vid, Dong2019FWGAN} and experiment with optical flow to improve temporal consistency between frames.
In particular, we use flow directional annotations to warp the previous generated frame into the current timestep.
A mask is then used to compose the warped frame with the originally synthesized result. 
We utilize a mask penalty, $\mathcal{L}_{F}$ weighted by $\lambda_{F}=\num{1e4}$ to constrain the learned flow mask. 
Our flow annotations are generated from FlowNet2 \cite{ilg2016flownet}.

During experiments, we find that flow causes issues with the general quality of generated images; specifically, flow introduces undesired artifacts to the video. The decreased quality is quantified by \figref{fig:flow}. For this reason, we omit flow in our reported best-performing result.
\section{Results} \label{sec:results}
We present the results of our bottom-up experiments. Our experiments show that the best-performing and most compatible model uses DensePose pose annotations, self-attention layers, and GeLU or ReLU activation function. We use this model to qualitatively and quantitatively compare with existing baselines.

\subsection{Evaluation Metrics}

In the field of virtual try-on, evaluation metrics are of high importance. Given the visual nature of virtual try-on, quantitative metrics do not tell the whole story. However, quantitative metrics are good at judging the overall quality of the output. 

\noindent\textbf{Structural Similarity Index (SSIM).}
SSIM is a perceptual metric that quantifies image quality caused by image processing. SSIM is often used in analyzing quality between videos. SSIM \cite{1292216} is defined by
\begin{equation}
    SSIM(x, y) = \dfrac{(2\mu_x\mu_y + c_1)(2\sigma_{xy} + c_2)}{(\mu_x^2 + \mu_y^2 + c_1)(\sigma_x^2 + \sigma_y^2 + c_2)}
\end{equation}
 where $\mu$ is the average, $\sigma^2$ is the variance, and $c_1$ and $c_2$ are constants used to stabilize the calculation. For our case, we use the multiscale SSIM metric, which applies SSIM across all input image channels, because it has been shown to perform better than vanilla SSIM.
 
\noindent\textbf{Peak Signal-to-Noise Ratio (PSNR).} 
PSNR is used to measure the quality of image reconstruction. Generally, this is used in the context of quality of image compression. We find that PSNR is a useful quantitative metric for judging the quality of a reconstructed image. PSNR is defined by
\begin{equation}
    PSNR(x, y) = 10 log(\dfrac{MAX^2_i}{MSE(x, y)})
\end{equation}
where MAX is largest pixel value in the input images, and MSE is defined as the mean square error between the $x$ and $y$.
Similar to SSIM, PSNR cannot be the standalone metric to judge the quality of try-on. That being said, PSNR is a sufficient approximation to human perception of image and video quality. Therefore, it is part of the comparison tools we use, for its general ability to judge and compare between good-quality and bad-quality videos.

\subsection{Quantitative Results}
\begin{figure}[h]
    \centering
    \includegraphics[width=8cm]{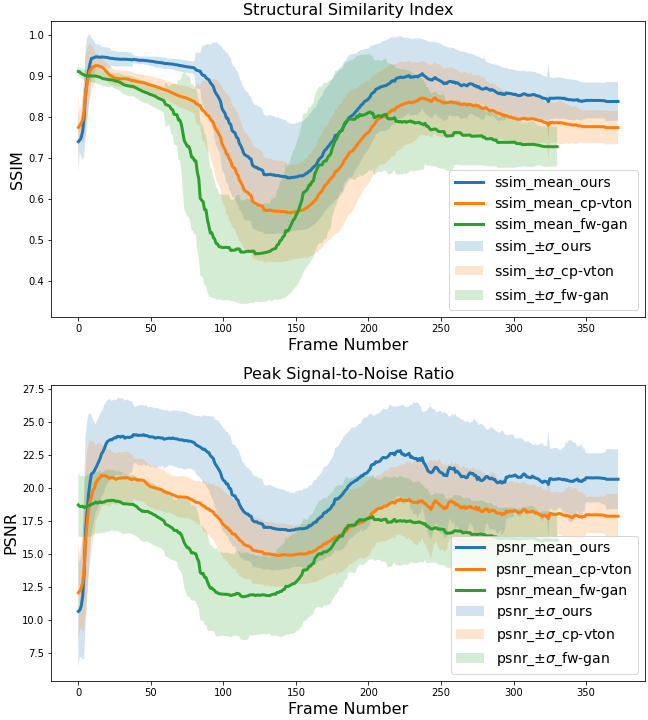}
    \caption{\textbf{Quantitative Comparison with Existing Try-On Methods.} We use SSIM and PSNR are our quantitative metrics to compare with \emph{CP-VTON} and \emph{FW-GAN} \cite{wang2018characteristicpreserving, Dong2019FWGAN}. We note, for completeness, that our problem statement differs from \emph{FW-GAN}. While that makes our approach more applicable for practical use, the quantitative comparison is incomplete.}
    \label{fig:comparison_across_modules}
\end{figure}

Through our bottom-up experiment structure, we determine the best method as per quantitative and qualitative analysis and determine the best-performing model as ShineOn. We used ShineOn methods and compared to existing try-on methods, CP-VTON and FW-GAN \cite{wang2018characteristicpreserving, Dong2019FWGAN}, 
to synthesize video virtual-try on results and compare. Using the SSIM and PSNR metrics, we calculate the mean and standard deviation at each frame across the three methods that we compare. Given that our problem statement is different from FW-GAN, \figref{fig:comparison_across_modules} reports these metrics and demonstrates that our method quantitatively outperforms existing try-on methods.

\subsection{Qualitative Results}

\newcolumntype{C}{>{\centering\arraybackslash}m{5.3em}}
\begin{table*}[] 
    \centering
    \small
    \begin{tabular}{*7{C}@{}}
        \toprule
        ID \& Frame Num. & Cloth (Input) & FW-GAN User Image (Input) & FW-GAN & VVT User Video (Input) & CP-VTON & ShineOn: DensePose, Attn, GELU \\ 
        \midrule
        4he21d00f-g11: frame 040
        & \includegraphics[width=5em]{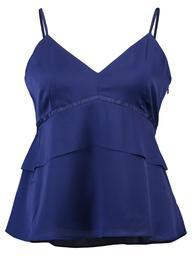}
        & \includegraphics[width=5.3em]{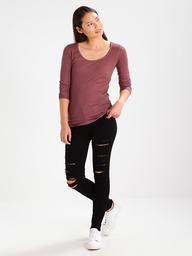}
        & \includegraphics[width=5.3em]{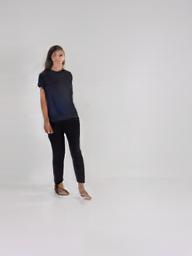}
        & \includegraphics[width=5.3em]{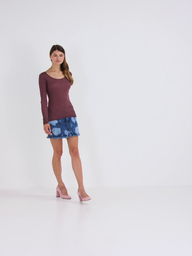}
        & \includegraphics[width=5.3em]{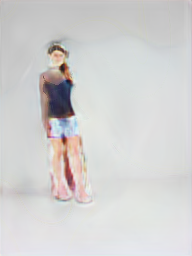}
        & \includegraphics[width=5.3em]{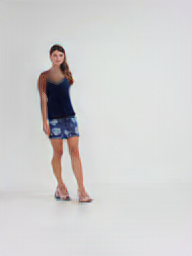}
        \\
        4he21d00f-g11: frame 125
        & \includegraphics[width=5em]{tryon/4he21d00f-g11/cloth.jpg}
        & \includegraphics[width=5.3em]{tryon/4he21d00f-g11/fwgan_user.png}
        & \includegraphics[width=5.3em]{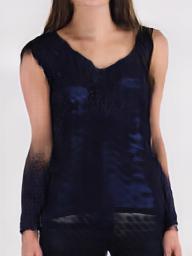}
        & \includegraphics[width=5.3em]{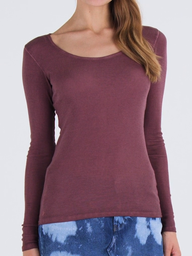}
        & \includegraphics[width=5.3em]{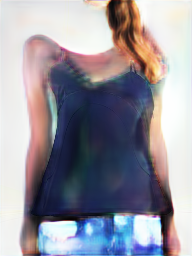}
        & \includegraphics[width=5.3em]{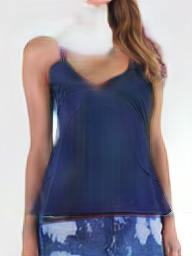}
        \\
        4he21d00f-k11: frame 040
        & \includegraphics[width=5em]{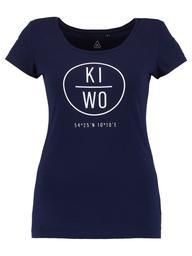}
        & \includegraphics[width=5.3em]{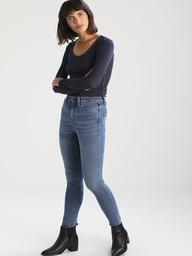}
        & \includegraphics[width=5.3em]{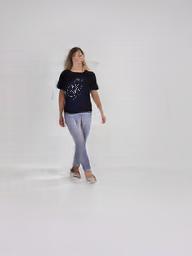}
        & \includegraphics[width=5.3em]{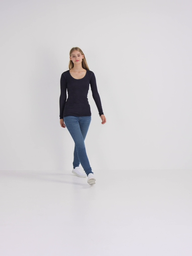}
        & \includegraphics[width=5.3em]{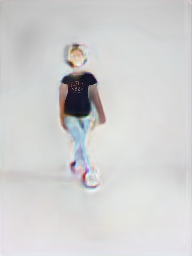}
        & \includegraphics[width=5.3em]{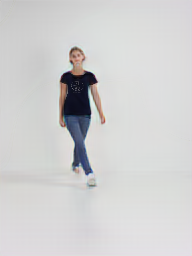}
        \\
        4he21d00f-k11: frame 125
        & \includegraphics[width=5em]{tryon/4he21d00f-k11/cloth.jpg}
        & \includegraphics[width=5.3em]{tryon/4he21d00f-k11/fwgan_user.png}
        & \includegraphics[width=5.3em]{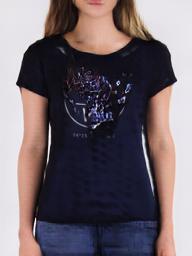}
        & \includegraphics[width=5.3em]{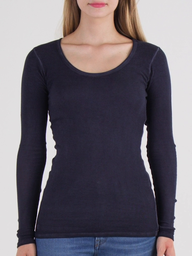}
        & \includegraphics[width=5.3em]{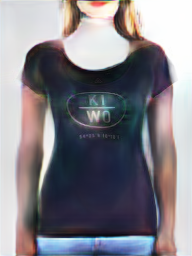}
        & \includegraphics[width=5.3em]{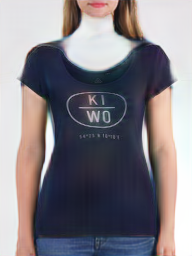}
        \\
        an621da9d-q11: frame 040
        & \includegraphics[width=5em]{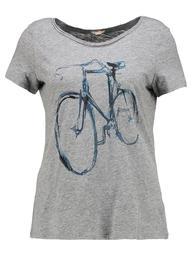}
        & \includegraphics[width=5.3em]{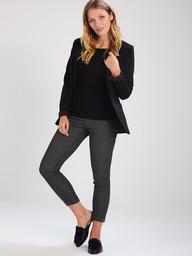}
        & \includegraphics[width=5.3em]{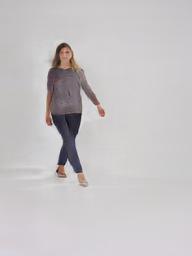}
        & \includegraphics[width=5.3em]{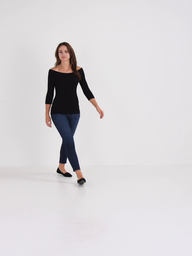}
        & \includegraphics[width=5.3em]{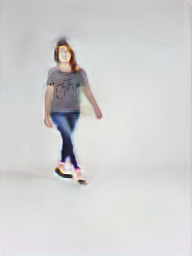}
        & \includegraphics[width=5.3em]{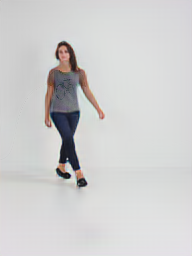}
        \\
        an621da9d-q11: frame 125
        & \includegraphics[width=5em]{tryon/an621da9d-q11/cloth.jpg}
        & \includegraphics[width=5.3em]{tryon/an621da9d-q11/fwgan_user.png}
        & \includegraphics[width=5.3em]{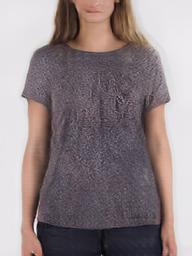}
        & \includegraphics[width=5.3em]{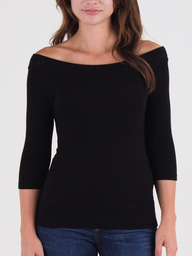}
        & \includegraphics[width=5.3em]{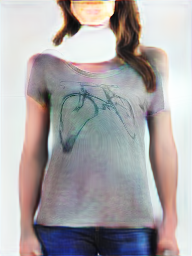}
        & \includegraphics[width=5.3em]{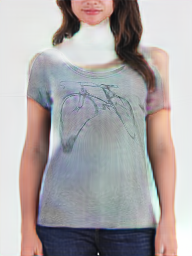}
        \\
        g1021d05g-k11: frame 040
        & \includegraphics[width=5em]{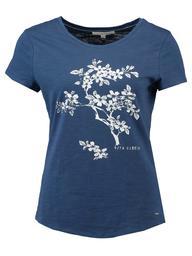}
        & \includegraphics[width=5.3em]{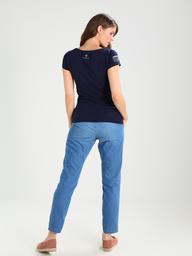}
        & \includegraphics[width=5.3em]{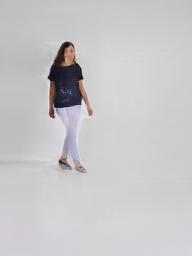}
        & \includegraphics[width=5.3em]{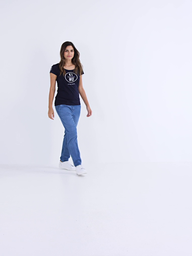}
        & \includegraphics[width=5.3em]{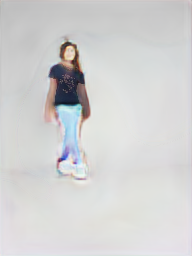}
        & \includegraphics[width=5.3em]{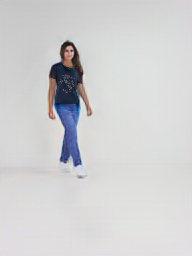}
        \\
        g1021d05g-k11: frame 150
        & \includegraphics[width=5em]{tryon/g1021d05g-k11/cloth.jpg}
        & \includegraphics[width=5.3em]{tryon/g1021d05g-k11/fwgan_user.png}
        & \includegraphics[width=5.3em]{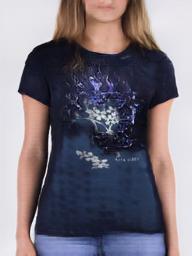}
        & \includegraphics[width=5.3em]{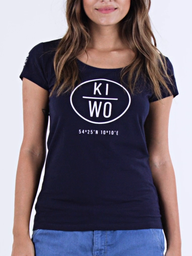}
        & \includegraphics[width=5.3em]{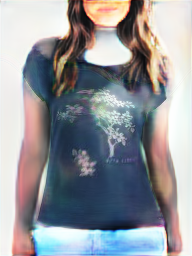}
        & \includegraphics[width=5.3em]{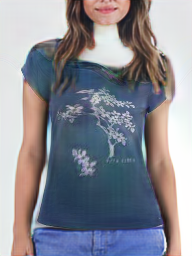}
        \\
        \bottomrule 
    \end{tabular}
    \caption{
        Qualitative try-on results comparing FW-GAN, CP-VTON (retrained on VVT), and the best ShineOn model that uses DensePose, Attention, and GeLU.
        As explained in Section \ref{subsec:problem_formulation}, FW-GAN reposes a still user image (column 3), while we directly use the user video frame (column 5).
        ShineOn better preserves user identity (face quality and other body parts), target cloth color, and target cloth texture design. However, ShineOn suffers from neck synthesis on zoomed views; we recommend exploring this issue in future work. 
    }
    \label{tab:tryon_methods_qual_compare}
\end{table*} 

Qualitative inspection and comparison are absolutely vital to judging the effectiveness of video virtual try-on.
Due to the importance of visual inspection in virtual try-on, qualitative criteria must be properly defined to evaluate the effectiveness of our presented methods.
These criteria are quality of body parts, quality of cloth synthesis, and temporal consistency.
We present visual comparisons in Table \ref{tab:tryon_methods_qual_compare}.
Our ShineOn methods outperform the CP-VTON baseline.
Our methods demonstrate strong body and face synthesis, better cloth transfer and texture, and temporal consistency.

Using the qualitative criteria to compare our methods with FW-GAN, we find that our method retains stronger face and body detail. It should be noted that our model's higher quality face detail arises from the differing problem formulations. Insofar as ShineOn and FW-GAN are video virtual try-on methods, this comparison stands. ShineOn preserves shirt designs more consistently to the product image, while the shirt designs in FW-GAN tend to be scattered. The weaknesses of our ShineOn are that it fails to generate video virtual try-on results with complete neck details. Furthermore, cloth try-on is increasingly misaligned during the frames where the camera is zoomed in on the person subject. The failure case of zoomed-in images is not specific to our network. FW-GAN also fails to effectively synthesize cloth for zoomed-in frames. Even though FW-GAN provides temporal smoothness due to temporal discriminators, it struggles with temporal consistency.

\section{Conclusion}
We methodically illuminate the effect of several design choices for practical video virtual try-on.
Our series of bottom-up experiments examine the outcomes of pose annotation choice, self-attention, activation functions, and optical flow.
The strengths and weaknesses of these design outcomes are compared to existing work.
Importantly, we release our code, hyperparameters, and model checkpoints to the public, not only to support experiment reproducibility but also as a framework for future works.

Within our experimental scope, we identify the design choices that result in the highest quality try-on.
Ordered by importance: 
(1) DensePose annotations improve face detail and decrease required memory and training time, (2) self-attention slightly benefits face and body feature quality, and (3) ReLU and GELU activations perform equally well, but not Swish nor Sine. 
We also identify that our design choice using optical flow improves temporal smoothness but introduced artifacts.
Our methodical approach to analyzing simple design choices results in significant improvement over the CP-VTON image try-on baseline. Though our best result struggles with neck details, it preserves user identity and shirt design better than FW-GAN.

For future work, we recommend investigating issues with synthesizing the neck.
We also suggest improving the cloth warping module, as it fails to handle 3D orientation and geometric information of the cloth (e.g. differentiating between inside and outside).
Future work may consider improving global temporal consistency instead of only local smoothness. 
Finally, as demonstrated here, we encourage reproducibility via methodical experiments supported by public code.


{\small
\bibliographystyle{ieee_fullname}
\bibliography{egbib}
}

\end{document}